\documentclass[letterpaper]{article} 
\usepackage{aaai24}  
\usepackage{times}  
\usepackage{helvet}  
\usepackage{courier}  
\usepackage[hyphens]{url}  
\usepackage{graphicx} 
\usepackage{natbib}  
\usepackage{caption} 
\usepackage{algorithm}
\usepackage{algorithmic}

\usepackage{newfloat}
\usepackage{listings}
\DeclareCaptionStyle{ruled}{labelfont=normalfont,labelsep=colon,strut=off} 
\floatstyle{ruled}
\newfloat{listing}{tb}{lst}{}
\floatname{listing}{Listing}
\usepackage{amsmath}
\usepackage{amssymb}
\usepackage{mathtools}
\usepackage{amsthm}

\usepackage{multirow}

\title{How important are specialized transforms in Neural Operators?}
\author {
    Ritam Majumdar \textsuperscript{\rm 1},
    Shirish Karande \textsuperscript{\rm 1},
    Lovekesh Vig \textsuperscript{\rm 1}
}
\affiliations {
    \textsuperscript{\rm 1}Tata Consultancy Services Research\\
    \{ritam.majumdar, shirish.karande,lovekesh.vig\}@tcs.com
}

\usepackage{bibentry}

\begin{document}

\maketitle

\begin{abstract}
Simulating physical systems using Partial Differential Equations (PDEs) has become an indispensible part of modern industrial process optimization. Traditionally, numerical solvers have been used to solve the associated PDEs, however recently Transform-based Neural Operators such as the Fourier Neural Operator and Wavelet Neural Operator have received a lot of attention for their potential to provide fast solutions for systems of PDEs. In this work, we investigate the importance of the transform layers to the reported success of transform based neural operators. In particular, we record the cost in terms of performance, if all the transform layers are replaced by learnable linear layers. Surprisingly, we observe that linear layers suffice to provide performance comparable to the best-known transform-based layers and seem to do so with a compute time advantage as well. We believe that this observation can have significant implications for future work on Neural Operators, and might point to other sources of efficiencies for these architectures.
\end{abstract}

\section{Introduction}

\begin{figure*}
\centering
\includegraphics[height=9cm, width=\textwidth]{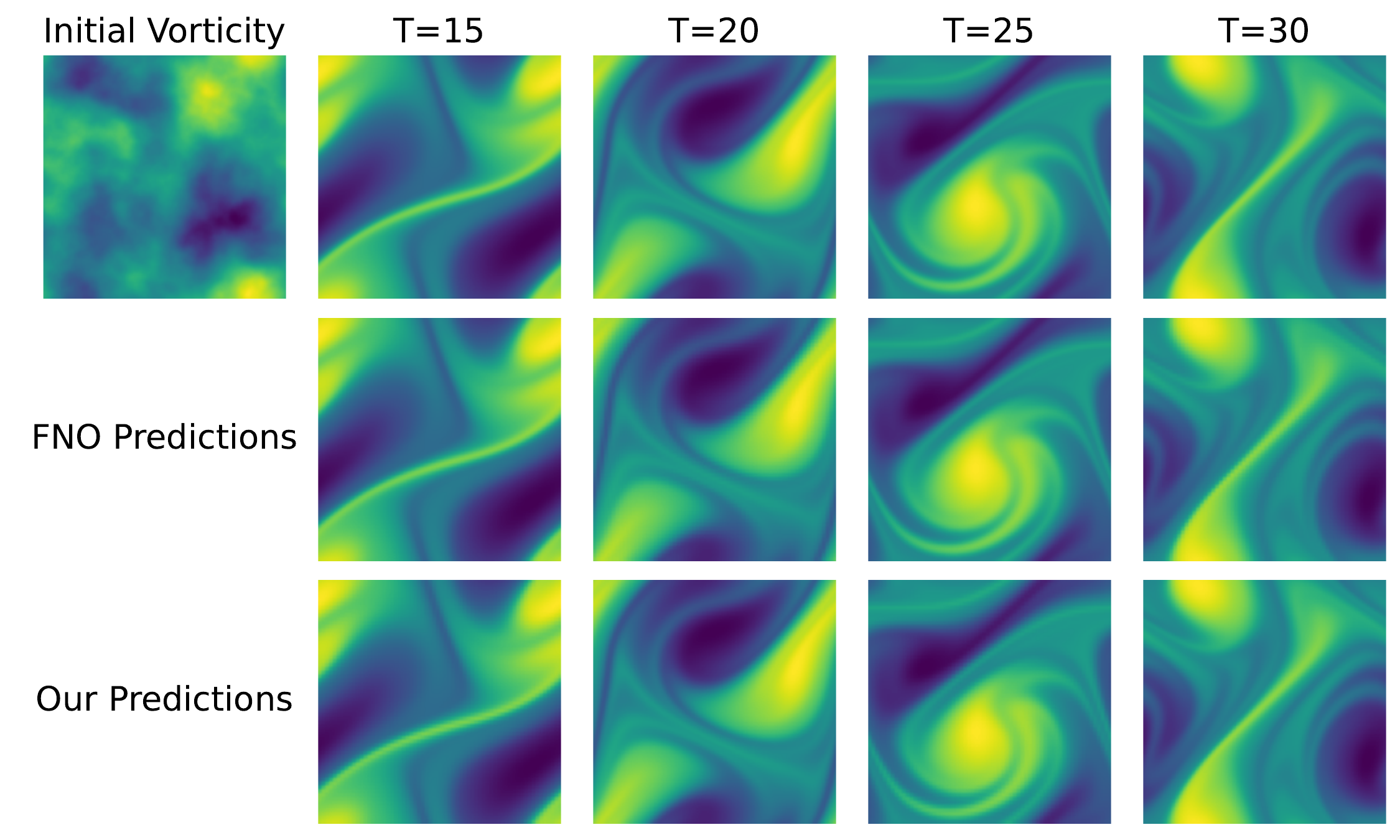}
\caption{Zero shot super-resolution comparison for Navier Stokes equation with viscocity $\nu=1e^{-4}$. First row represents ground-truth, while second and third rows represent predictions by FNO and parameterized linear transform based neural operator respectively. The neural operators are trained on $64\times64\times20$ dataset and evaluated on $256\times256\times80$ dataset.}
\label{fig:Title_plot}    
\end{figure*}
The prediction and analysis of physical systems with the help of computational forward simulations have emerged as a crucial instrument in numerous industrial domains. These physical systems, constrained by an array of Partial Differential Equations (PDEs) along with initial and boundary values, call for precise and scalable simulation methodologies. Among the diverse range of techniques available, transform-based Neural Operators such as Fourier Neural Operator (FNO) \cite{FNO} and Wavelet Neural Operator (WNO) \cite{WNO} have attracted considerable attention for their ability to deliver rapid, scale-free simulations.

One of the salient features of these transform-based operators is their ability to transform the input data into a domain where the data's inherent features become more apparent, thus making it easier to analyze and perform computations. Traditionally, the efficacy of such transformations is highly reliant on the intrinsic nature of the data, leading to an optimal choice of transform becoming a significant determinant of the overall performance of the system. However, despite the growing interest and success of these methods, a comprehensive investigation into the necessity of these sophisticated transforms, and their effect on system performance is lacking. Tremendous human effort is involved in the study of characteristics of the industrial data and making the appropriate choice of transform.

The universal approximation theorem \cite{UAT} states that a neural network with a single hidden layer containing a finite number of neurons can approximate continuous functions on compact subsets, given the activation function is a non-constant, bounded, and monotonically increasing continuous function. This includes specialized transforms like Fourier, Laplace, and Wavelet transforms. It has been demonstrated, Neural Networks can successfully compute Discrete Fourier Transforms \cite{DFTNN}. As specific transforms lead to distinct transformation of the PDE-induced input data, the choice of transform becomes critical to the generalization capabilities of the neural operator.  While there is significant effort involved in designing the correct transform for given input data and PDEs, most considered transforms are linear in nature, raising the question: Are these well-known pre-defined  transforms essential, or could they be replaced by a parameterized linear layer which can learn an adaptive  transform for the specific problem under consideration and deliver comparable or better performance? This study probes this pertinent question and explores the performance implications when replacing all transform layers in the network with basic learnable linear layers.

We hypothesize that learnable linear transformations suffice in terms of generalization and computational efficacy. The initial results of our exploration are surprising, pointing towards linear layers exhibiting performance parity with the best-known transform-based operators, seemingly with a computational advantage. Such a revelation prompts further questioning of the underlying significance of transform-based operators in the Neural Operator realm. This study serves as an exploration of this notion, providing empirical evidence that challenges the status quo of existing Neural Operator models. The primary objective is to delve into the trade-offs between using a pre-defined linear transform and using a parameterized linear transform which adapts to the PDE problem in-hand. We analyze the performance factor and the computational cost as criterion for our trade-offs. By doing this, we aim to shed light on the potential of learnable linear transformations and their capability to generalize over varying PDE systems of different complexity, without having to handcraft specialized transforms and save human effort.

Our contributions are summarized as follows: 1) We replace Fourier (Wavelet) transform with a parameterized learnable linear transform. 2) Learnable transform based NOs outperform WNOs and perform on-par with FNOs including long-temporal Navier Stokes and Kolmogorov flow even on super-resolution settings. 3) Learnable-transforms are quicker to train than FNOs and WNOs due to fewer number of parameters as parameters in FNO/WNO belong to the complex domain, while we define parameters of the linear transform entirely in the real domain.

\section{Related Work}
\label{Related Work}

Traditional approaches to forward simulation of Partial Differential Equations (PDEs) primarily involve numerical methods such as Finite Difference, Finite Volume, and Finite Element Methods. These approaches discretize the problem domain into a grid or mesh and approximate the derivatives in the PDEs using this discretized representation. However, these methods are computationally demanding as they often require small time steps for stability, particularly in high-dimensional spaces. Furthermore, they face challenges in handling complex geometries, multiple scales, and non-linearities inherent in many physical systems. In the recent past, Neural network based techniques like Physics-Informed Neural Networks (PINNs), \cite{PINN}, Deep Galerkin Methods (DGM) \cite{DGM}, Deep BSDE solvers \cite{DeepBSDE} have been proposed as surrogates for solutions of PDEs. DGM is a neural network-based approach for solving PDEs that formulates the solution as a continuous optimization problem, handling high-dimensional problems more effectively than traditional methods. Deep BSDE Solvers solve high-dimensional PDEs by representing them as backward stochastic differential equations. These methods however, don't generalize to changing initial-boundary conditions and require retraining for every unique initial-boundary conditions and parameterized PDE coefficients.

Operator-based methods circumvent around this issue by aiming to learn a mapping or an operator, that transforms a given set of input conditions (parameterized initial-boundary conditions) to the corresponding solutions. Given a new-instance of a PDE system, instead of developing solutions from scratch, a pre-trained operator can simply infer the solutions in real time, reducing computational cost. Some widely-used operators for learning systems of parameterized PDEs are DeepONets \cite{DeepONets}, Fourier Neural Operators (FNO) \cite{FNO}, Wavelet Neural Operators (WNO) \cite{WNO}, Laplace Neural Operators (LNO) \cite{LNO}, Multiwavelet Neural Operator \cite{gupta2021multiwaveletbased} just to name a few. DeepONets, rooted in the universal approximation theorem, consist of a branch network processing function parameters and a trunk network handling function variables. The final output, integration of the two network outputs, effectively approximates complex operators. Special linear-transforms like Fourier, Wavelet and Laplace have been incorporated into Neural transforms as tools for feature extraction and data compression to improve the representation-capacity of Neural Operators. FNO consists of a Fourier transform to convert the input data into Fourier space, followed by multiple layers of 1D convolutions and nonlinearities to learn the mapping, and then an inverse Fourier transform to convert the output back to the original space. WNO takes Wavelet transforms instead of Fourier transforms and is better equipped to handle non-periodic and irregular domains due to the superiority of wavelets in time-frequency localization. LNO uses the Laplace transform to decompose the input space and can handle non-periodic signals, take transient responses into account, and converge exponentially. The above neural operators have also been extended to their Physics-informed variations, Physics-informed Neural Operators \cite{PINO}, Physics-informed Wavelet Neural Operators, \cite{PIWNO} to circumvent the limitations of purely data-driven methods.

\section{Methodology}
\label{Methodology}

\begin{figure*}
    \centering
\includegraphics[height=12cm, width=\textwidth]{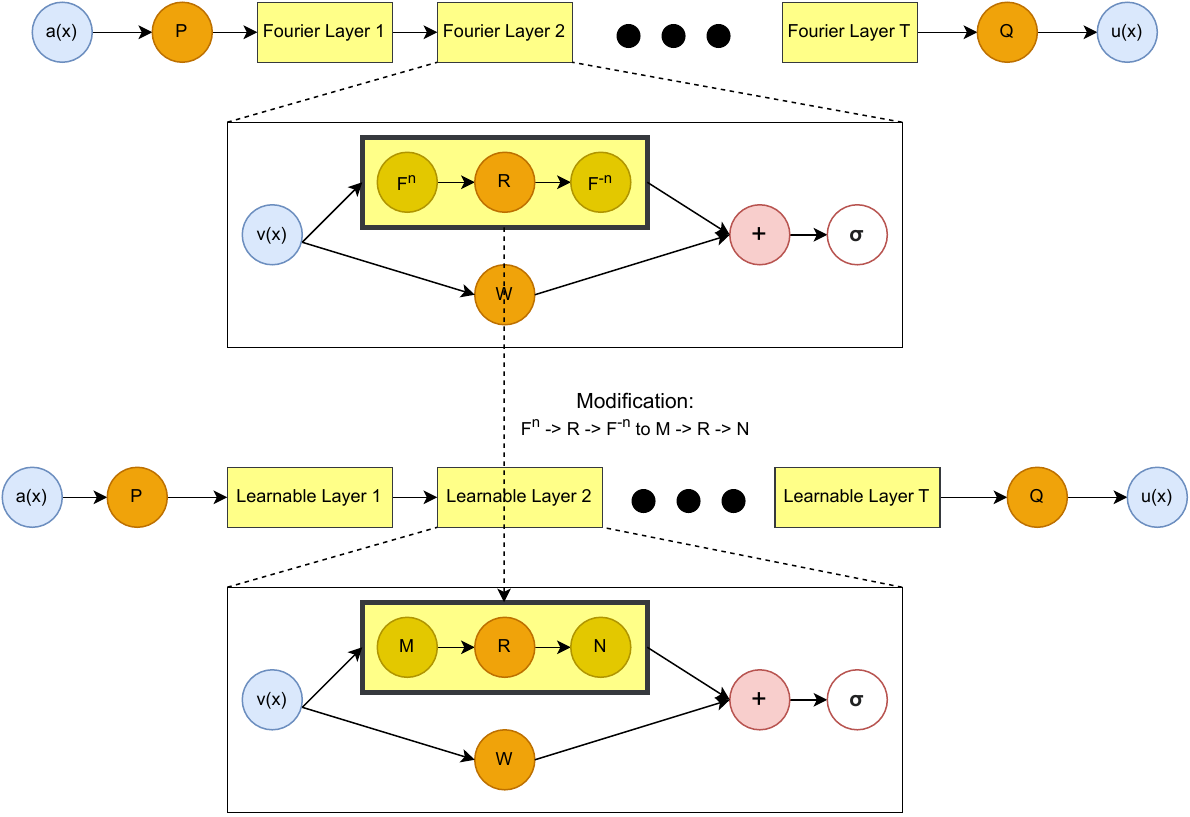}
    \caption{Model architecture of Neural Operator. Top: Architecture of Fourier Neural Operator as proposed in \cite{FNO}. Bottom: The final architecture of the learnable-transform based Neural Operator. We just replace the central solid block from $F^n\rightarrow R \rightarrow F^{-n}$ to $M\rightarrow R \rightarrow N$. Here $F^n$ represents the n-dimensional fourier transform, $F^{-n}$ represents n-dimensional inverse-fourier transform respectively, while $M$ and $N$ refer to learnable forward transformation and learnable inverse transformation respectively.}
    \label{fig:Image}
\end{figure*}

We describe the architecture of our Neural Operator in Figure \ref{fig:Image}. The architecture resembles the original architectures of FNO and WNO, the only difference being, we replace the one-dimensional Fourier (Wavelet) and Inverse Fourier (Wavelet) transforms with learnable linear transforms $M$ and $N$ respectively. Our objective is to learn a operator $G$ parameterized by $\theta$ mapping $G_{\theta}:A\rightarrow U$, wherein $A = A(D; R^{d_a})$ and $U = U(D; R^{d_u} )$ are input and output Banach spaces and $D \subset R^d$ be a bounded, open set. $a(x)\in R^{d_a}$ and $u(x)\in R^{d_u}$ refer to the input and output signals sampled i.i.d from $A$ and $U$ respectively. 

The neural operator, similar to \cite{FNO} is formulated as an iterative architecture $v_0 \rightarrow
v_1 \rightarrow ... \rightarrow v_T$ where $v_j$ with $j \in [0,T-1]$ is a sequence of functions each taking values in $R^{d_v}$. $R^{d_v}$ is the dimension of the middle projection layers. As shown in Figure \ref{fig:Image}, the input $a\in A$ is first lifted to a higher dimensional representation 
\begin{equation}
    v_0(x) = P(a(x))
\end{equation} by the local transformation $P: R^{d_a} \rightarrow R^{d_v}$ which is parameterized by a shallow neural network. Then we apply several iterations of updates $v_t \rightarrow v_{t+1}$ defined as the composition of a non-local integral operator K and a local, nonlinear activation function $\sigma$. The projection Q transforms the projection back to the dimension of the output space $Q: R^{d_v} \rightarrow R^{d_u}$, while the output $u(x)$ is defined as 
\begin{equation}
    u(x) = Q(v_T(x))
\end{equation} The update $v_t \rightarrow v_{t+1}$ is defined as:
\begin{equation}
    v_{t+1}(x)=\sigma(W.v_t(x)+K(a;\phi)v_t(x))\\ 
\end{equation}
\begin{equation}
    K(a;\phi)v_t(x) = \int_{D} \kappa(x,y,a(x),a(y); \phi)v_t(y)dy\\
\end{equation}
\begin{equation}
  \kappa(x,y,a(x),a(y); \phi)v_t(y) = N(R\cdot M(v_t(y)))\\
\end{equation} 
Here $\kappa(\phi): R^{2(d+d_a)}\rightarrow R^{d_vXd_v}$ is a kernel integral operator represented by a neural network parameterized by $\phi$. $W: R^{d_v} \rightarrow R^{d_v}$ is a linear transformation, and $\sigma$ is a
non-linear activation function. 

An n-dimensional Fourier transform can be broken down into n 1D-Fourier Transforms along each dimension \cite{tolimieri2012mathematics}. Given an $n$-dimensional tensor of shape $R^{d_1} \times R^{d_2} \times \cdots \times R^{d_n}$, the $n$-dimensional Fourier transform $M$ of the tensor is defined as:
\begin{equation}
M(R^{d_1} \times \cdots \times R^{d_n}) = F(R^{d_1}) \times \cdots \times F(R^{d_n})
\end{equation}
where $F$ refers to the one-dimensional Fourier transform. The resulting tensor is of the shape $C^{k_1} \times C^{k_2} \times \cdots \times C^{k_n}$ where $k_1, k_2, \ldots, k_n$ are the chosen number of Fourier modes in the low-frequency regime, and $C$ refers to the complex domain. We replace the one-dimensional Fourier transform $F$ with a learnable linear layer $L_f$ and $L_b$ where $L_f$ stands for linear transform in forward space and $L_b$ refers to linear transform in the inverse space.
We define our $n$-dimensional linear transformation $M$ as follows:
\begin{equation}
M(R^{d_1} \times \cdots \times R^{d_n}) = L_{f_1}(R^{d_1}) \times \cdots \times L_{f_n}(R^{d_n})
\end{equation}
where $L_{f_i}$ where $i \in [1, n]$ represents learnable linear transforms (parameterized matrices of shape $R^{d_i}\times R^{k_i}$) along each dimension. In figure \ref{fig:Image}, R refers to the tensor multiplication in the transformed space. In the Fourier space, $R$ takes in $C^{dv} \times C^{k_1} \times \cdots \times C^{k_n}$ dimensional tensor as input, takes an element wise tensor multiplication with a parameterized tensor of shape $C^{dv^2} \times C^{k_1} \times \cdots \times C^{k_n}$ to output a $C^{dv} \times C^{k_1} \times \cdots \times C^{k_n}$ dimensional tensor. In our case, we stay in the real domain, and our input and output tensors are of the shape $R^{dv} \times R^{k_1} \times \cdots \times R^{k_n}$, while the parameterized tensor is of the shape $R^{dv^2} \times R^{k_1} \times \cdots \times R^{k_n}$.

The n-dimensional inverse transform $N$ is defined as
\begin{equation}
N(C^{k_1} \times \cdots \times C^{k_n}) = F^{-1}(C^{k_1}) \times \cdots \times F^{-1}(C^{k_n})
\end{equation}
where $F^{-1}$ refers to the one-dimensional inverse Fourier transform. The resulting tensor is of the shape $R^{d_1} \times R^{d_2} \times \cdots \times R^{d_n}$, the original dimensions of the tensor. In our case, we replace $F^{-1}$ with $L_b$ (parameterized matrices of shape $R^{k_i}\times R^{d_i}$) and define the inverse transform $N$ to be
\begin{equation}
N(R^{k_1} \times \cdots \times R^{k_n}) = L_{b_1}(R^{k_1}) \times \cdots \times L_{b_n}(R^{k_n})  
\end{equation}
The M-R-N operation is repeated iteratively for T blocks, before being projected to the output space $u(x)$ using projection $Q$ defined in Eqn 2. In Fourier Neural Operators, $M$ and $N$ are exact inverse operations, i.e. Fourier transform and Inverse fourier transform. In learnable transforms however, we don't constrain $M\cdot N=I$ and keep them flexible.   

\begin{figure*}
    \centering
\includegraphics[height=6cm, width=\textwidth]{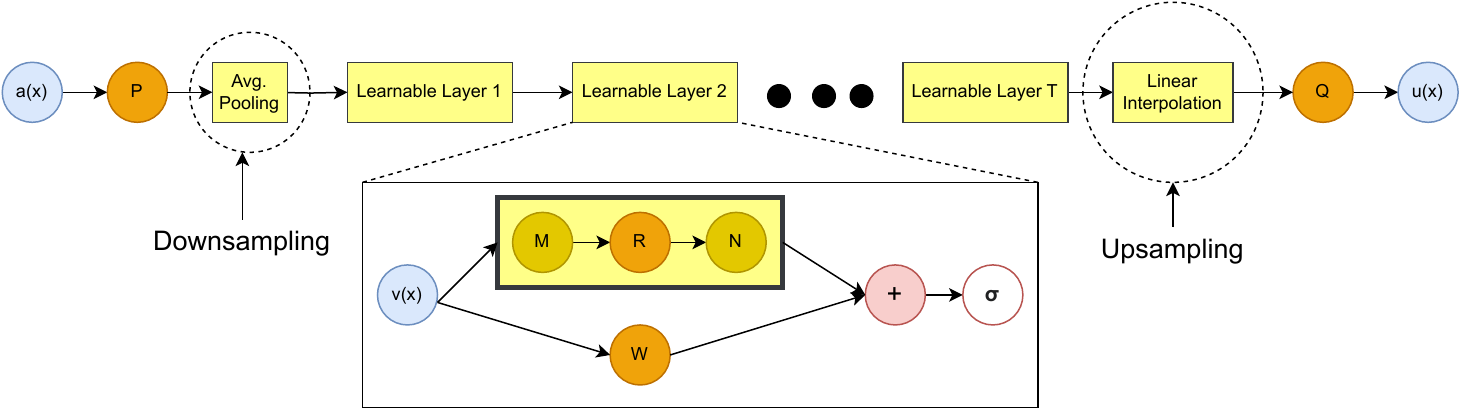}
    \caption{Architecture of our Neural network at test-time, to handle super-resolution. We add two blocks, Average pooling before applying the first learnable layer to downsample and Linear-interpolation block after applying the final learnable layer. Both these blocks have no-parameters.}
    \label{fig:LearnableNOTest}
\end{figure*}

Fourier Neural Operators handle super-resolution by considering the lower-frequency modes of the transformed tensor, irrespective of the resolution of input tensor. In Figure \ref{fig:LearnableNOTest}, we demonstrate how we handle super-resolution at test-time. Before passing through the first learnable layer, we downsample the tensor to the dimensions used at train-time using average pooling. We apply all the learned transformation blocks, then upsample the tensor to the desired resolution using linear interpolation before passing it  through the projection layer $Q$. For one-dimensional cases, we use "nearest" interpolation, while in two-dimensional and three-dimensional cases, we use "bilinear" and "trilinear" interpolations respectively.

\section{PDE information}
In order to make a fair comparison, we consider the same class of examples used in the original papers of \cite{FNO,WNO}. 

\subsection{Burgers' Equation} We consider the 1D Burgers' equation, which is a non-linear PDE with various applications, including modeling the flow of a viscous fluid. The 1D Burgers' equation takes the form:
$$
\begin{aligned}
\partial_t u(x, t)+\partial_x\left(u^2(x, t) / 2\right) & =\nu \partial_{x x} u(x, t), & & t \in(0,1] \\
u(x, 0) & =u_0(x), & & x \in(0,1)
\end{aligned}
$$
where $u_0$ is the initial condition and $\nu = 1e^{-3}$ is the viscosity coefficient. We aim to learn the operator mapping the initial condition $u(x,0) $ to solution at time $t=1$: $u(x,1)$. 

\subsection{Wave advection equation} The wave advection equation is a hyperbolic PDE and primarily describes the solution of a scalar under some known velocity field. The advection equation with periodic boundary condition is given by,

$$
\begin{aligned}
\partial_t u(x,t)+v \partial_x u(x,t)=0,\quad\quad\quad x,t\in(0,1)\times(0,1] \\
u(x-\pi,t)=u(x+\pi,t), \quad\quad\quad  x,t\in(0,1)\times(0,1]\\
u(x, 0)=h_{\left\{c-\frac{\omega}{2}, c+\frac{\omega}{2}\right\}}+\sqrt{\max \left(h^2-(a(x-c))^2, 0\right)}
\end{aligned}
$$

 $v\in\mathbb{R}>0$ represents the speed of the flow. $u(x,0)$ represents the initial condition, the variables $\omega$ and $h$ represent the width and height of the square wave, respectively and the wave is centered at $x=c$. For $v=1$, the solution to the advection equation is given as, $u(x, t)=u_0(x-t)$. The objective is to learn the mapping $u_0(x) \mapsto u(x, t)$, to a later time $t$. 

\subsection{2D Darcy Flow} 2D Darcy Flow is a linear second-order elliptic PDE. We consider a steady-state flow in a unit box, given by:
$$
\begin{aligned}
-\nabla \cdot(a(x) \nabla u(x)) & =f(x), & & x \in(0,1)^2 \\
u(x) & =0, & & x \in \partial(0,1)^2
\end{aligned}
$$
with a Dirichlet boundary where $a$ is the diffusion coefficient and $f=1$ is the forcing function. The objective is to learn the operator mapping the diffusion coefficient $a(x)$ to the solution $u(x)$. 

\subsection{2D Darcy flow equation with a notch in triangular domain} This example is a special case of the 2D-Darcy problem in the triangular domain with a notch in the flow. The boundary conditions for the triangular domain are generated using the following Gaussian process (GP), with the kernel $\mathcal{K}$,

$$
\begin{aligned}
u(x)\sim\mathrm{GP}\left(0, \mathcal{K}\left(x, x^{\prime}\right)\right) \quad\quad\quad\quad && x,x^{\prime}\in[0,1] \\
\mathcal{K}\left(x, x^{\prime}\right)=exp(-\frac{(x-x^{\prime})^2}{2l^2} )\quad\quad && l=0.2
\end{aligned}
$$

The permeability and the forcing function $f(x, y)$ is 0.1 and -1 respectively. The objective is to learn the operator mapping the boundary conditions to the pressure field, given by, $\left.u(x, y)\right|_{\partial \omega} \mapsto u(x, y)$. 

\subsection{Navier-Stokes Equation} The two dimensional temporally varying Navier-Stokes equation for a viscous, incompressible fluid on the unit torus is given by:
$$
\begin{aligned}
\partial_t w+u\cdot \nabla  =\nu \Delta w+f(x), \quad\quad\quad x,t\in (0,1)^2 \times (0,T) \\
\nabla \cdot u=0,\quad\quad\quad\quad\quad\quad\quad x,t\in (0,1)^2 \times (0,T) \\
w(x,0)=w_0(x),\quad\quad\quad\quad\quad\quad\quad\quad\quad x\in(0,1)^2\\
f(x)=0.1\left(\sin \left(2 \pi\left(x_1+x_2\right)\right)+\cos \left(2 \pi\left(x_1+x_2\right)\right)\right)\quad\quad
\end{aligned}
$$

where $w=\nabla\times u$ is the vorticity, $w_0$ is the initial vorticity, $\nu \in \mathbb{R}_{+}$is the viscosity coefficient, and $f(x)$ is the forcing function. The objective is to learn the operator mapping the vorticity up to time 10 to the vorticity from time 10 to some later time. 

\subsection{Kolmogorov Flow} The high-frequency Kolmogorov flow is governed by the 2D Navier-Stokes equation, defined by,

$$
\begin{aligned}
\partial_t w+u\cdot \nabla  =\nu \Delta w+f(x), \quad\quad\quad x,t\in (0,2\pi)^2 \times (0,T) \\
\nabla \cdot u=0,\quad\quad\quad\quad\quad\quad\quad x,t\in (0,2\pi)^2 \times (0,T) \\
w(x,0)=w_0(x),\quad\quad\quad\quad\quad\quad\quad\quad\quad x\in(0,2\pi)^2\\
f(x)=\sin \left(n x_2\right) \hat{x}_1\quad\quad\quad\quad\quad\quad\quad\quad\quad\quad\quad\quad\quad
\end{aligned}
$$

Here $x_1$ is the unit vector in the X-direction. The objective is to learn the evolution operator mapping the next time-step from the previous time-step.

\section{Training and Hyperparameter Details}

In order to have a fair comparison and benchmark in an unbiased manner, we consider the original PDE examples and datasets used in the original papers of \cite{FNO, WNO}, same set of hyperparameters (Same number of layers, same spatial and temporal grid dimensions of PDE examples, same transform dimensions, optimizers, schedulers, batch sizes of examples, etc). We construct our neural operator by stacking four integral operator layers with ReLU as the activation function. We use an Adam optimizer for 500 epochs with an initial learning rate of $1e^{-3}$ that is halved every 100 epochs. All experiments were conducted on Nvidia P100 GPU with 16 GB GPU Memory and 1.32 GHz GPU Memory clock using Pytorch framework. In Burger's, Wave-Advection, 2D-Darcy Rectangular and Navier-Stokes with $\nu=\{1e^{-3},1e^{-5}\}$, we consider 1000 training examples and 200 test-examples,Navier-Stokes with $\nu={1e^{-4}}$, we consider 10000 training examples and 200 test-examples, while in Darcy Triangular domain with notch we consider 1900 and 100 train and test examples respectively. Our hidden dimensions of linear transforms are same as the number of modes in Fourier transform used in \cite{FNO}. For one-dimensional examples, our final transform-dimension is 64, while for two-dimensional and three-dimensional examples, our transform-dimension is 32. Across all examples, we use \% relative L2-error as our metric.

\section{Results and Discussion}
\label{Results-and-discussion}

In this section, we discuss the results of our experiments. We divide our analysis into two sections, Table \ref{tab:Results1} consists of the lower-dimensional PDEs, while Table \ref{tab:Results2} consists of higher-dimensional PDEs. Across all examples in Tables \ref{tab:Results1},\ref{tab:Results2} the experiments are trained on the lower resolution and evaluated on same and higher resolutions. In Table \ref{tab:Results2}, the property column refers to viscosity in Navier-Stokes and Reynold's number (Re) in Kolmogorov flow. The complexity of the problem is increased by the decrease in viscosity in Navier-Stokes and higher Re in Kolmogorov flow, which leads to turbulent behavior. 

\subsection{Lower-dimensional PDEs}
From Table \ref{tab:Results1}, we observe our Neural Operator with a learnable linear transform performs competitively against the best-performing architecture between FNO and WNO. FNO struggles on the example of Wave-Advection as it fails to directly map one-dimensional initial condition $u_0$ to two-dimensional output, and simply repeating $u_0$ multiple times to form a two-dimensional input leads to poor training loss \cite{Lu_2022}. Fourier transform struggle to handle irregular geometry, we use a variant of FNO, dgFNO+ introduced in \cite{Lu_2022} to handle Darcy-flow in triangular domain with a notch. Our neural linear transform is capable of learning the transform suitable for irregular geometries, indicated by the superior performance over FNOs and WNOs. Numerically, learnable linear-transform based Neural operators are within 0.001 to 0.002 points (outperforms Darcy-triangular-notch by 0.05 points) of the best-performing transform between FNO and WNO across all examples. Similar to FNOs, the errors of our Neural Operator don't increase drastically on testing over higher resolution grids. We use average sampling to downsample and apply forward transform across all examples, while we use "nearest" interpolation and "bilinear" interpolation for one-dimensional and two-dimensional examples respectively. Using simple linear interpolations to scale from lower to higher resolutions suffices to keep the errors minimal, and we speculate that using more sophisticated interpolation techniques may further provide better solutions in higher resolution regime.

\begin{table}
    \centering
    \begin{tabular}{c c c c c}
        PDE & Resolution & Ours & FNO & WNO\\ 
        \hline
            & 1024 & 0.173 & \textbf{0.171} & 0.20 \\
        Burger's & 2048 & \textbf{0.175} & \textbf{0.175} & 0.22  \\
            & 4096 & \textbf{0.177} & \textbf{0.177} & 0.22  \\
            \hline
        \multirow{2}{*}{Wave Advection} & $40\times40$ & 0.61	& 45.14 & \textbf{0.59}\\
        & $80\times80$ & \textbf{0.64}	& 48.22 & \textbf{0.64}\\
            \hline
         & $106\times106$ & \textbf{0.271} & 0.272 & 0.34\\
      Darcy-flow & $211\times211$ & \textbf{0.273} & \textbf{0.273} & 0.35\\
        & $421\times421$ & \textbf{0.274} & 0.275 & 0.35\\
        \hline
      \multirow{2}{*}{Darcy-Notch} & $50\times50$& \textbf{0.67} & $0.72^\textbf{*}$ & 0.71\\
      &$100\times100$  & \textbf{0.69} & $0.75^\textbf{*}$ & 0.84\\
        \hline
    \end{tabular}
    \caption{Test \% relative L2-error for lower dimensional PDEs}
    \label{tab:Results1}
\end{table}

\subsection{Higher dimensional PDEs}
\begin{table}
    \centering
    \begin{tabular}{c c c c  c c}
        PDE & Property & Resolution& Ours & FNO & WNO\\
        \hline
        &&64x64&\textbf{0.83}&{0.84}&1.78\\
	&$1e^{-3}$&128x128&\textbf{0.85}&{0.86}&1.88\\
        &&256x256&\textbf{0.88}&{0.89}&2.02\\
        &&64x64&{7.07}&\textbf{7.04}&9.39\\
	NS&$1e^{-4}$&128x128&{7.19}&\textbf{7.16}&9.66\\
        &&256x256&{7.34}&\textbf{7.32}&9.94\\
    &&64x64&15.27&\textbf{14.85}&18.57\\
	&$1e^{-5}$&128x128&{15.49}&\textbf{15.29}&20.25\\
        &&256x256&{15.71}&\textbf{15.58}&23.41\\
        \hline
        \multirow{6}{*}{K'rov}&\multirow{2}{*}{100}&64x64&\textbf{3.96}&{4.04}&5.53\\
        &&128x128&\textbf{4.16}&{4.24}&5.87\\
        &\multirow{2}{*}{400}&64x64&9.26&\textbf{9.15}&10.73\\
        &&128x128&9.48&\textbf{9.34}&11.25\\
        &\multirow{2}{*}{500}&64x64&13.36&\textbf{13.35}&15.54\\
        &&128x128&\textbf{13.51}&\textbf{13.51}&16.06\\
        \hline
        \hline
        &&64x64&\textbf{0.96}&{0.98}&2.05\\
	&$1e^{-3}$&128x128&\textbf{1.01}&{1.02}&2.14\\
        &&256x256&\textbf{1.05}&{1.07}&2.25\\
        &&64x64&{7.22}&\textbf{7.17}&9.89\\
	NS&$1e^{-4}$&128x128&{7.30}&\textbf{7.26}&10.36\\
        &&256x256&{7.37}&\textbf{7.34}&10.57\\
    &&64x64&15.32&\textbf{15.04}&20.21\\
	&$1e^{-5}$&128x128&{15.56}&\textbf{15.26}&21.13\\
        &&256x256&{15.70}&\textbf{15.51}&26.34\\
        \hline
        \multirow{6}{*}{K'rov}&\multirow{2}{*}{100}&64x64&\textbf{4.02}&{4.07}&5.75\\
        &&128x128&\textbf{4.25}&{4.29}&5.99\\
        &\multirow{2}{*}{400}&64x64&9.51&\textbf{9.45}&11.05\\
        &&128x128&9.74&\textbf{9.62}&11.44\\
        &\multirow{2}{*}{500}&64x64&13.56&\textbf{13.52}&16.01\\ &&128x128&{13.71}&\textbf{13.69}&16.25\\
        \hline

    \end{tabular}
    \caption{Test \% relative L2-error for higher dimensional PDEs. NS refers to Navier-Stokes temporal flow, while K'rov refers to Kolmogorov flow. In case of NS, property refers to the viscosity of the flow, while in K'rov, it refers to the Reynold's number. The first 2 sections represent the results trained using a 3D operator, while the last 2 sections represent the results trained using a 2D operator + LSTM for the temporal component.}
    \label{tab:Results2}
\end{table}

Table \ref{tab:Results2} represents the results on higher-dimensional PDEs, the more difficult examples in our study. In Navier-Stokes, the spatial and temporal components are modeled using a 3D-Neural Operator. The training resolution is set to $64\times64$ for training. We observe, across all examples, taking a learnable linear transform performs comparably against Fourier transform and beats Wavelet transforms across all examples. The learnable transforms are within 0.42 points against FNOs, while being the best performing architecture on $\nu=1e^{-3}$. In the Kolmogorov flow examples, learnable transforms based NOs are within 0.1 points against FNOs. This observation remains consistent even when we use a 2D-Operator to model the spatial component and an LSTM to model the temporal component, although 3D-operators perform better against a 2D+LSTM operator across all examples. In order to test super-resolution capabilities of 3D examples, we downsample the input by average pooling before passing it through the first operator block, and upsample them using trilinear interpolation to recover the original size. We notice, like FNOs, the error almost remains in the same range while testing for super-resolution, indicating the effectiveness of a simple trilinear interpolation. Additionally, the increase in error when predicting a higher resolution is similar to that of FNOs. Navier-Stokes: (0.03 for $\nu=1e^{-3}$, 0.15 for $\nu=1e^{-4}$ and 0.22 for $\nu=1e^{-5}$). Similar observations hold for Kolmogorov flow as well. Wavelet transforms have a drastic increase in error when tested with higher resolution samples, especially when the task complexity increases. Thus, parameterized transforms have the ability to scale to the complexity of the PDE and can adapt to learn a transform best suited to the PDE problem under consideration. 

\subsection{Training-time and \# parameter analysis}
We refer to the training time per epoch for each of the operators in Table \ref{tab:Results3}. We train all the operators for 500 epochs (the original hyperparameter considered in FNO and WNO papers) to keep the benchmarking consistent. Learnable linear transforms train faster than FNOs and WNOs across all examples. We investigate the reasons behind this, and make a comparsion of the parameter count of differentiating blocks of each operator in Table \ref{tab:Parametersizecomparison}. All notations in Table \ref{tab:Parametersizecomparison} are consistent with those defined in the methodology section of the paper. $M, N$ refers to the forward and inverse transforms, while $R$ refers to the tensor multiplication. $n$ refers to number of dimensions, while $d_i,k_i$ refer to input and output dimensions before and after computing a forward transform, respectively. While FNO/WNO have no extra parameters while undergoing the forward, inverse transforms, the tensor multiplication $R$ takes place in the complex space, contributing to an additional ${d_v}^2\prod_{i=1}^{n}{k_i}$ parameters as against learnable linear-layer based transforms. In all scenarios, FNO/WNO architectures have an additional ${d_v}^2\prod_{i=1}^{n}{k_i}-2\sum_{i=1}^{n}{d_ik_i}$ parameters per block. The training time of a neural network architecture is proportional to the number of parameters. Since linear-transform based Neural Operators end up having a lower number of parameters, their training time is faster as compared to FNO/WNO.

\begin{table}
    \centering
    \begin{tabular}{c c c c}
        PDE & Ours & FNO & WNO\\
        \hline
        1D-Burgers & 2.52 & 2.75 & 5.58\\
        Advection & 2.72 & 2.86 & 6.03\\
        2D-Darcy & 13.55 & 13.84 & 38.34\\
        Navier-Stokes (2D+time)& 67.41 &130 & 221\\
        Navier-Stokes (3D)& 22.14 & 46.33 & 196\\
        Kolmogorov & 16.35 & 19.56& 47.75\\
        \hline
    \end{tabular}
    \caption{Training times of operators (per epoch) in seconds}
    \label{tab:Results3}
\end{table}

\begin{table}
    \centering
    \begin{tabular}{c c c}
         & FNO/WNO & Ours \\
         \hline
         \vspace{2pt}
         $M$&0&$\sum_{i=1}^{n}{d_ik_i}$\\
         \vspace{2pt}
         $R$&$2{d_v}^2\prod_{i=1}^{n}{k_i}$&${d_v}^2\prod_{i=1}^{n}{k_i}$\\
         \vspace{2pt}
         $N$&0&$\sum_{i=1}^{n}{d_ik_i}$\\
         \vspace{2pt}
        Total&$2{d_v}^2\prod_{i=1}^{n}{k_i}$&${d_v}^2\prod_{i=1}^{n}{k_i}+2\sum_{i=1}^{n}{d_ik_i}$\\
        \hline
    \end{tabular}
    \caption{Parameter size comparison per block}
    \label{tab:Parametersizecomparison}
\end{table}

\section{Conclusion}
\label{Conclusion}

We replace specialized transforms like Fourier and Wavelet in Neural Operators with a learnable linear transform. The learnable linear transforms generalize well and perform competitively on a wide class of examples like low-viscous Navier Stokes and high Re Kolmogorov flow. Additionally, learnable linear transforms generalize on Wave-Advection and irregular domain Darcy flows which traditional Fourier transforms fail to do and is computationally faster than Wavelet transform-based operators. Thus, a parameterized linear layer can replace specialized transforms and saves human effort for Neural Operator architecture design. In future, we seek to investigate the learnt transforms and their resemblance with known transforms. Additionally, theoretical studies on convergence rates and error-bounds for parameterized linear layer-based Neural Operators remain.

\section{Broader impact}

The findings of this work hold the potential to streamline computational simulations across various industries, from engineering and meteorology to physics, by utilizing learnable linear transformations instead of pre-defined transform-based Neural Operators. This may foster faster, more accessible and more generalizable simulations, expanding their applicability even to higher complexity entities.

\bibliography{aaai24}

\end{document}